\documentclass[11pt]{article}

\usepackage[final]{acl}

\usepackage{times}
\usepackage{latexsym}
\usepackage{multirow}
\usepackage{enumitem}

\usepackage{amsmath}
\usepackage[inkscapelatex=false]{svg}

\usepackage[T1]{fontenc}
\usepackage[utf8]{inputenc}

\usepackage{microtype}

\usepackage{inconsolata}

\usepackage{graphicx}

\title{Some Emotions Run Deeper: Layer-wise Probing and Causal Intervention in Large Language Models}

\author{
  Tian Fang\textsuperscript{1}, Gaël Guibon\textsuperscript{1,2}, Davide Buscaldi\textsuperscript{1} \\
  \textsuperscript{1}Universit\'e Sorbonne Paris Nord, CNRS, Laboratoire d’Informatique de Paris Nord, LIPN\\
  \textsuperscript{2}LORIA, CNRS, Universit\'e de Lorraine\\
  \texttt{\{fang,guibon,buscaldi\}@lipn.fr}
}

\begin{document}
\maketitle
\begin{abstract}
Emotion is expressed in text along a wide spectrum, from surface
lexical cues to inferences entangled with content. Most layer-wise
analyses of emotion in LLMs use a single corpus, leaving open whether
the depth at which emotion becomes accessible is a property of the model
or also of the text source. We investigate this across three datasets
spanning different degrees of explicitness and contextualization in
emotion expression (Twitter posts, Reddit comments, and autobiographical
narratives) and eight 1B--9B open-weight LLMs from the Llama, Qwen, and
Granite families. We combine layer-wise probing with offline feature scaling and online forward interventions, transfer analyses, and an early-exit classifier.
We find that (i) the best probing layer shifts systematically across
corpora, from input-adjacent layers to over half model depth, and this
ordering persists after matching label-by-length-bin distributions; (ii)across the evaluated settings, forward-pass interventions on probe-selected bands reduce
test accuracy by 5--6 points more than same-width random bands
($q < 0.01$); (iii) selected bands transfer across datasets and emotion
categories, suggesting partially shared affective information rather
than strictly per-emotion substrates; and (iv) probe-selected early-exit
representations outperform full-depth exits by $6.9$ percentage points on average.
\end{abstract}

\section{Introduction}
\label{sec:intro}
Large language models are increasingly used in affect-sensitive settings,
including emotion classification, emotional support, and broader tests of
emotional intelligence
\citep{sabour-etal-2024-emobench,wang2023emotional,kang-etal-2024-large}.
Although prior evaluations show that LLMs can recognize and respond to
emotional content at the output level, less is known about how affective
information is organized internally: whether it becomes accessible at a
fixed depth in decoder-only LLMs, or varies with the kind of text being
processed.

This question matters because emotion is expressed differently across
text sources. In short social-media posts, affect often appears through
surface cues such as lexical markers, emotion hashtags, emoji/emoticons,
punctuation, or conventional phrases
\citep{10.1111/coin.12024,
mohammad-bravo-marquez-2017-emotion,https://doi.org/10.1111/j.1468-2885.2010.01362.x}.
In longer comments and autobiographical narratives, it may instead be
conveyed implicitly through situations, appraisals, causal structure, or
pragmatic inference
\citep{balahur-etal-2011-detecting,hofmann-etal-2020-appraisal,
troiano-etal-2023-dimensional,scherer1994evidence}.

Layer-wise probing is a natural tool for locating where emotion becomes
accessible in LLM representations
\citep{alain2017understanding,belinkov-2022-probing}. Prior work shows
that linguistic, semantic, and factual information peak at different
transformer depths
\citep{tenney-etal-2019-bert,rogers-etal-2020-primer,
geva-etal-2023-dissecting}, and that sentiment and emotion are most
detectable in early-to-mid Llama layers
\citep{palma-etal-2025-llamas}. Whether this affective pattern
generalizes across model families and text sources remains unclear.
Because decodability alone does not imply functional relevance
\citep{hewitt-liang-2019-designing,belinkov-2022-probing}, we pair
probing with intervention-based perturbations and early-exit readout
\citep{Fan2020Reducing,10.1016/j.csl.2022.101429,
mcgrath2023hydraeffectemergentselfrepair,xin-etal-2020-deebert}.

\begin{figure*}[t]
    \centering
    \includegraphics[width=\textwidth]{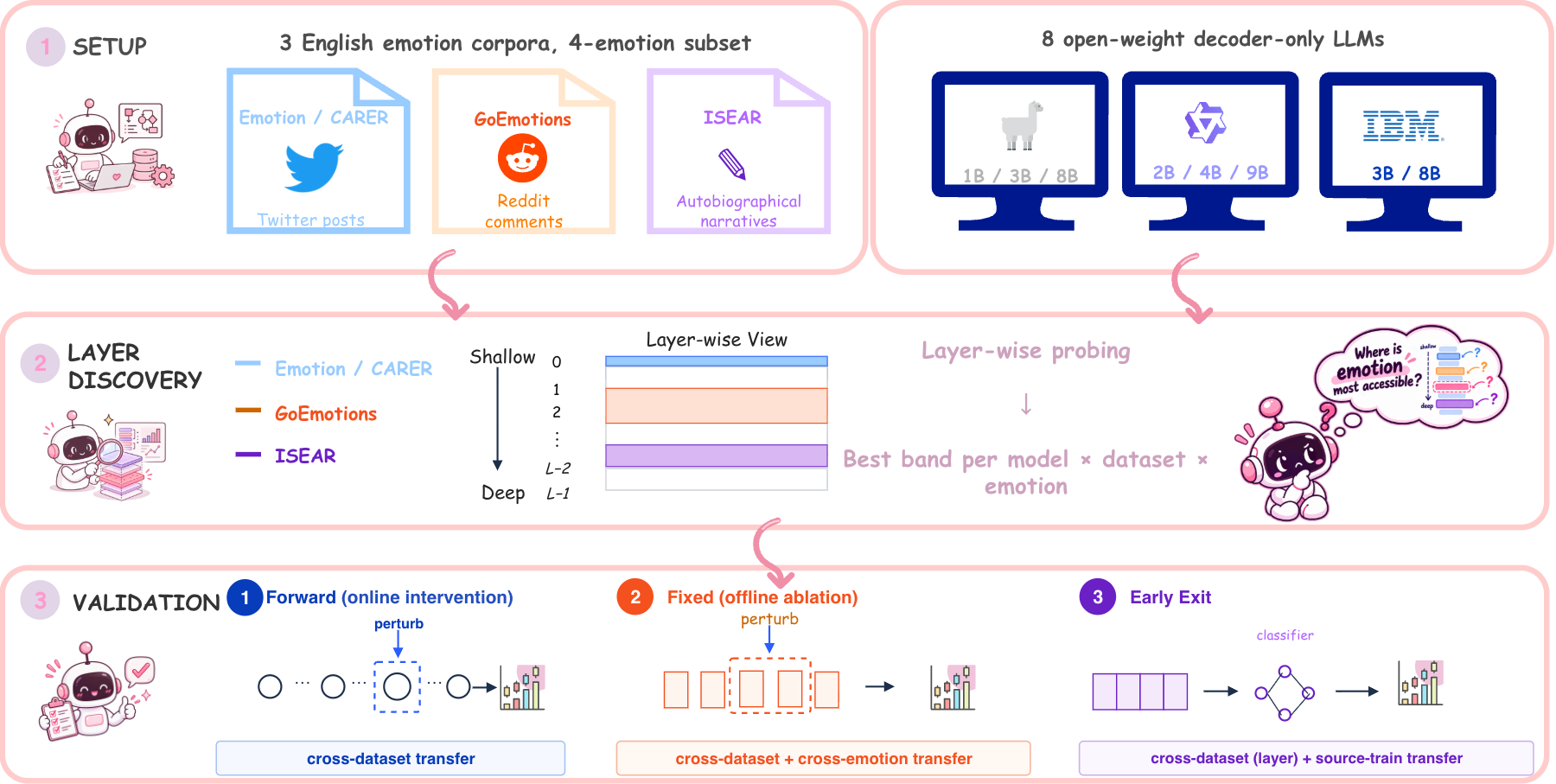}
    \caption{
Overview of our experimental pipeline. We localize emotion-sensitive layer bands with
layer-wise probing, then evaluate them through online intervention,
offline feature scaling, transfer analyses, and early-exit
classification.
}
    \label{fig:pipeline}
\end{figure*}

We study where emotion-sensitive information becomes accessible in
decoder-only LLMs, and whether probe-selected layers are merely
diagnostic or also useful for intervention and readout. We analyze eight
open-weight LLMs from three families (Llama, Qwen, and Granite; 1B--9B
parameters) on three stylistically different English emotion corpora:
\textit{Emotion}/CARER, \textit{GoEmotions}, and \textit{ISEAR}. We
restrict them to the shared labels fear, joy, anger, and sadness, use
frozen representations with fixed deterministic pooling and linear
probes, and evaluate selected layer bands through online intervention,
offline feature scaling, transfer analyses, and early-exit
classification. Figure~\ref{fig:pipeline} summarizes the pipeline.

Our main findings are fourfold:
\begin{itemize}[noitemsep, topsep=0pt]
    \item Emotion depth varies systematically with dataset source:
    Emotion/CARER peaks near input-adjacent layers, GoEmotions is
    intermediate and diffuse, and ISEAR peaks much deeper.
    \item Probe-selected bands are causally engaged in the
    probe-mediated readout, as online interventions disrupt prediction
    more than same-width unrelated controls.
    \item Transfer depends on the evaluation setting, with stronger
    sharing under intervention than under early-exit readout.
    \item Probe-selected bands support compact readout: across the evaluated
settings, selected early exits outperform, on average, full-depth exits
and control exits based on random, low-sensitivity, or transferred layers
under the same linear-probe protocol.
\end{itemize}

Together, these results show that affective information is localized enough to be causally disrupted and efficiently exploited, but its depth and generalizability are not universal: they depend on what kind of text is being processed, which model family is used, and how the representation is decoded.

\section{Related Work}
  Our work draws on three lines of prior research: emotion
  understanding in large language models,
  layer-wise probing of model representations, 
  and layer-level interventions together with their use for efficient
  inference. 

\noindent \textbf{Emotion Understanding in Language Models.}
  \label{sec:emotion}
 Discrete emotion categorization is a long-standing NLP task, grounded
  in psychological theories of basic emotions
  \citep{Ekman01051992, plutchik1980general}. A growing body of work
  has evaluated large language models on emotion classification across
  narrative, social-media, and dialogue domains
  \citep[e.g.,][]{wang2023emotional, sabour-etal-2024-emobench}, generally
  reporting strong end-to-end performance. This literature characterizes what an LLM predicts rather than how the prediction arises from internal representations. We complement these benchmark-driven results with a representation-level probing analysis of decoder-only LLMs, treating internal hidden states as the unit of study.

\noindent \textbf{Layer-wise Probing.}
    \label{sec:probing}
Probing classifiers \citep{alain2017understanding, belinkov-2022-probing} have become a standard methodology for analyzing information encoded in intermediate representations of neural language models. \citet{hewitt-liang-2019-designing} argue that probes
  should be as simple as possible, typically linear, so that findings
  reflect representation quality rather than probe capacity. Layer-wise analyses on BERT-style models have consistently found that lower layers tend to capture surface and syntactic features while higher layers encode more abstract semantic information \citep{tenney-etal-2019-bert, rogers-etal-2020-primer}. Related forms of depth-structured processing have also been observed for factual recall in decoder-only language models \citep{geva-etal-2023-dissecting}. Beyond linguistic and
  factual content, probing has also been used to localize numerical
  reasoning \citep{wallace-etal-2019-nlp}, spatial and temporal knowledge
  \citep{gurnee2024language}, and high-level concept directions such
  as truthfulness in instruction-tuned models \citep{marks2024the}. 
The most directly related work, \citet{palma-etal-2025-llamas},
  probes sentiment and emotion in the Llama family to identify an
  effective probe-and-pooling combination, reporting that affective
  information is most detectable in the early-to-mid layers. Our focus
  is on how emotion is organized across multiple model families,
  datasets, and individual emotion categories, validated through
  causal interventions.

\noindent \textbf{Layer-level Interventions and Efficient Inference.}
    \label{sec:interventions}
  While probing identifies what information is decodable from
  representations, it does not establish whether that information plays
  a causal role in model behavior. A complementary line of work tests
  causal contribution by intervening on internal representations.
  Layer-level interventions, which ablate, drop, or perturb entire
  layers and measure the behavioral consequence
  \citep{Fan2020Reducing, 10.1016/j.csl.2022.101429, mcgrath2023hydraeffectemergentselfrepair}, are
  particularly suited to questions about whether information at a given
  depth is causally required for a downstream task.

  A second line of work leverages the layer-organized structure of
  transformers for inference-time efficiency. Early-exit methods
  \citep{schwartz-etal-2020-right, xin-etal-2020-deebert,schuster2022confident} learn policies that halt computation at an
  intermediate layer when an input is sufficiently easy, trading
  accuracy for compute. In the emotion setting,
  \citet{palma-etal-2025-llamas} apply this idea to Llama via
  SENTRILLAMA, which truncates the model at a single probe-identified
  layer. We extend this direction by comparing layer bands of varying
  width to test whether a contiguous multi-layer band outperforms
  single-layer selection, and by evaluating cross-dataset transfer of
  selected bands.

\section{Preliminaries}
\noindent \textbf{Datasets.}
We evaluate on three publicly available emotion-classification
corpora loaded from HuggingFace:
\textbf{Emotion}~\citep{saravia-etal-2018-carer}, Twitter posts labeled with
six emotions; \textbf{ISEAR}~\citep{scherer1994evidence},
autobiographical narratives elicited via emotion-recall questionnaires
across seven emotional categories; and \textbf{GoEmotions}
\citep{demszky-etal-2020-goemotions}, Reddit comments labeled with 27
fine-grained emotions plus neutral. From each corpus, we retain the
shared target labels fear, joy, anger, and sadness, and discard examples
shorter than six whitespace tokens. For GoEmotions, we additionally keep
only single-label examples whose sole label is one of the four target
emotions. We then construct stratified splits with seed 42: 80/10/10 for
Emotion and ISEAR, and 70/15/15 for the resulting smaller GoEmotions subset.
Emotion is capped at 1{,}500 examples per emotion before splitting.
Table~\ref{tab:dataset_sizes} gives per-split sizes; additional dataset
statistics and corpus-cue diagnostics are reported in
Appendix~\ref{app:dataset_stats}.
  \begin{table}[t]
  \centering
  \small
  \begin{tabular}{lrrrr}
  \hline
  Dataset & Train & Val & Test & Total \\
  \hline
  Emotion (CARER) & 4{,}800 & 600 & 600 & 6{,}000 \\
  \hline
  ISEAR           & 3{,}234 & 402 & 408 & 4{,}044 \\
  \hline
  GoEmotions      & 2{,}147 & 459 & 463 & 3{,}069 \\
  \hline
  \end{tabular}
  \caption{Dataset sizes after restriction to the four-emotion subset
  and stratified splitting (80/10/10 for Emotion and ISEAR;
  70/15/15 for GoEmotions).}
  \label{tab:dataset_sizes}
  \end{table}

  \noindent \textbf{Models.}
  We probe eight open-weight decoder-only large language models from
  HuggingFace across three families:
  \begin{itemize}[topsep=2pt, itemsep=0pt, leftmargin=*]
      \item \textbf{Llama}: Llama-3.2-1B-Instruct, Llama-3.2-3B-Instruct,
            Llama-3.1-8B-Instruct~\citep{grattafiori2024llama3herdmodels}
      \item \textbf{Qwen}: Qwen-3.5-2B, Qwen-3.5-4B,
            Qwen-3.5-9B~\citep{qwen3.5}
      \item \textbf{Granite}: Granite-4.1-3B,
            Granite-4.1-8B~\citep{granite41}
  \end{itemize}

\noindent \textbf{Pooling and Probing.}
Our work does not aim to identify the optimal pooling function or probe
classifier for emotion probing, a question recently studied at length by
\citet{palma-etal-2025-llamas}. For the main experiments, we restrict
pooling to fixed, deterministic operations applied identically across
models, layers, and datasets. We use a validation-only, greedy two-stage
pilot to select a common pooling--probe configuration. First, using a
linear SVM on Llama-3.2-3B-Instruct and the Emotion validation split, we
compare last-token, mean, and \texttt{concat(mean, max, min)} pooling
across the four target emotions; a smaller ablation additionally confirms
that the concatenated representation outperforms max or min pooling alone.
Second, with \texttt{concat} fixed, we compare linear SVM, logistic
regression, and MLP probes. Logistic regression provides the strongest
overall combination of validation performance, simplicity, and stability,
although the MLP is marginally better in some individual cells. The
selected pooling and probe choices are also favored in a cross-family
sanity check for fear on Qwen-3.5-4B.

As reported in Appendix~\ref{app:pooling_probe_pilot}, a learned
single-head attention-pooling baseline can perform strongly at favorable
shallow-layer settings, but is substantially more sensitive to layer,
random seed, and optimization hyperparameters, with some deeper-layer runs
collapsing to zero validation F1. We therefore retain the non-parametric
pooling configuration. After this validation-only pilot, we fix a
\texttt{StandardScaler}--\texttt{LogisticRegression} pipeline
($C=1.0$, \texttt{max\_iter}=5000,
\texttt{random\_state}=42) on top of
\texttt{concat(mean, max, min)} pooling for all main analyses.

\section{Where and How do emotions reside in LLMs?}
\label{sec:layerwise}

Before intervention, we first localize where emotion information is linearly decodable in frozen LLM representations. For each dataset, model, target emotion, and layer, we train a probe on the training split and evaluate it on validation only. Because models have different depths, we report normalized depth, $(\ell+1)/L$, where $\ell$ is the zero-indexed layer and $L$ is the total number of layers.
This analysis localizes where emotion information is linearly decodable; causal relevance is assessed later through fixed and online interventions.

\begin{figure*}[t]
  \centering
  \includegraphics[width=\textwidth]{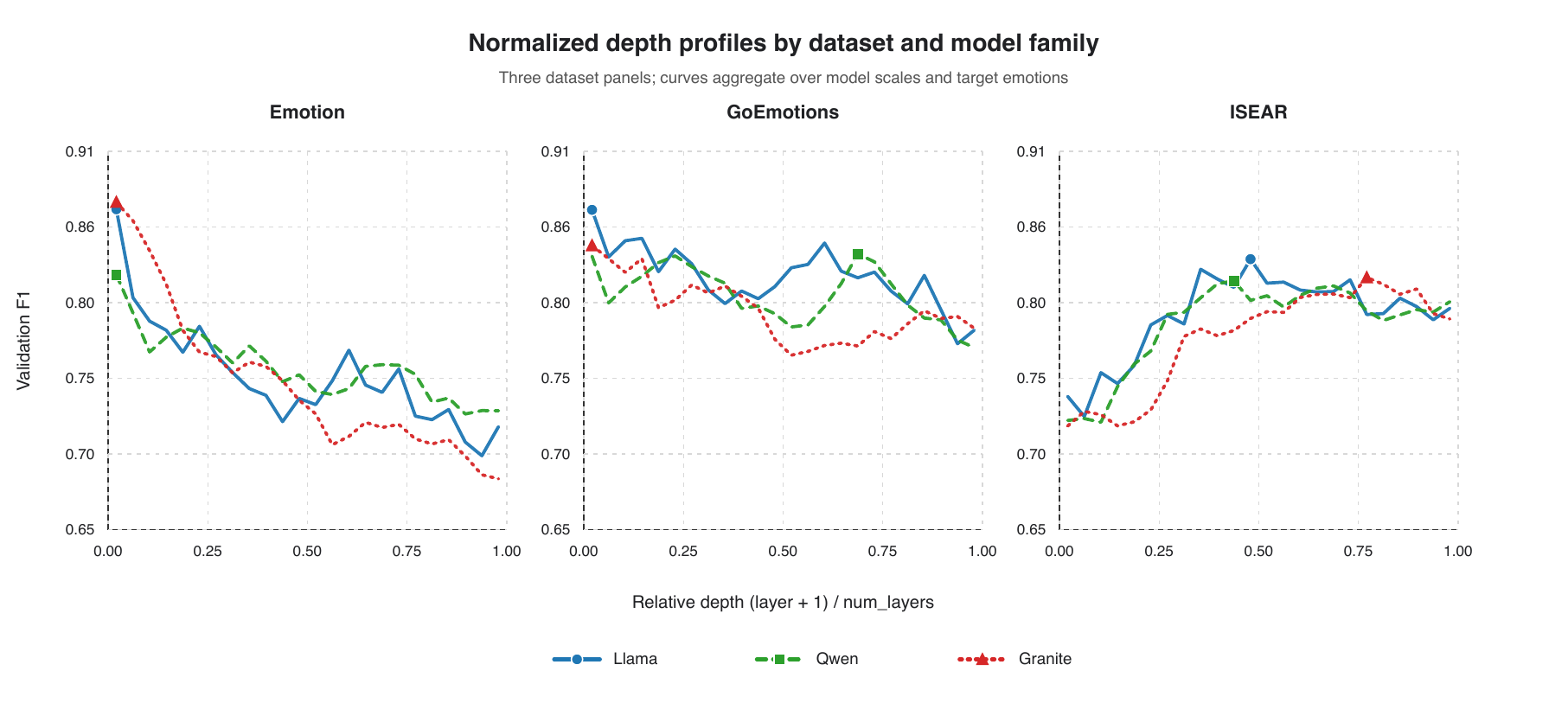}
  \caption{
  Layer-wise emotion decodability over normalized depth. Each panel corresponds to a dataset, and each curve aggregates validation F1 over model scales and target emotions within a model family. Across families, Emotion peaks early, GoEmotions is broader and intermediate, and ISEAR peaks later.
  }
  \label{fig:layerwise-family-dataset}
\end{figure*}

\noindent \textbf{Dataset source style shifts emotion-cue depth.}
Figure~\ref{fig:layerwise-family-dataset} shows normalized-depth profiles across datasets and model families. The main pattern is dataset-driven. Emotion peaks very early, with mean best relative depth $0.066$ and median depth $0.036$; GoEmotions is more intermediate and diffuse, with mean best depth $0.219$; and ISEAR peaks much later, with mean and median depths $0.590$ and $0.598$. This ordering persists on length-matched subsets,
with mean best relative depths of $0.131$, $0.288$, and $0.642$ for
Emotion, GoEmotions, and ISEAR, respectively. Thus, short social-media-style posts expose emotion cues early, whereas autobiographical ISEAR narratives require deeper processing.

\begin{figure*}[t]
  \centering
  \includegraphics[width=.82\textwidth]{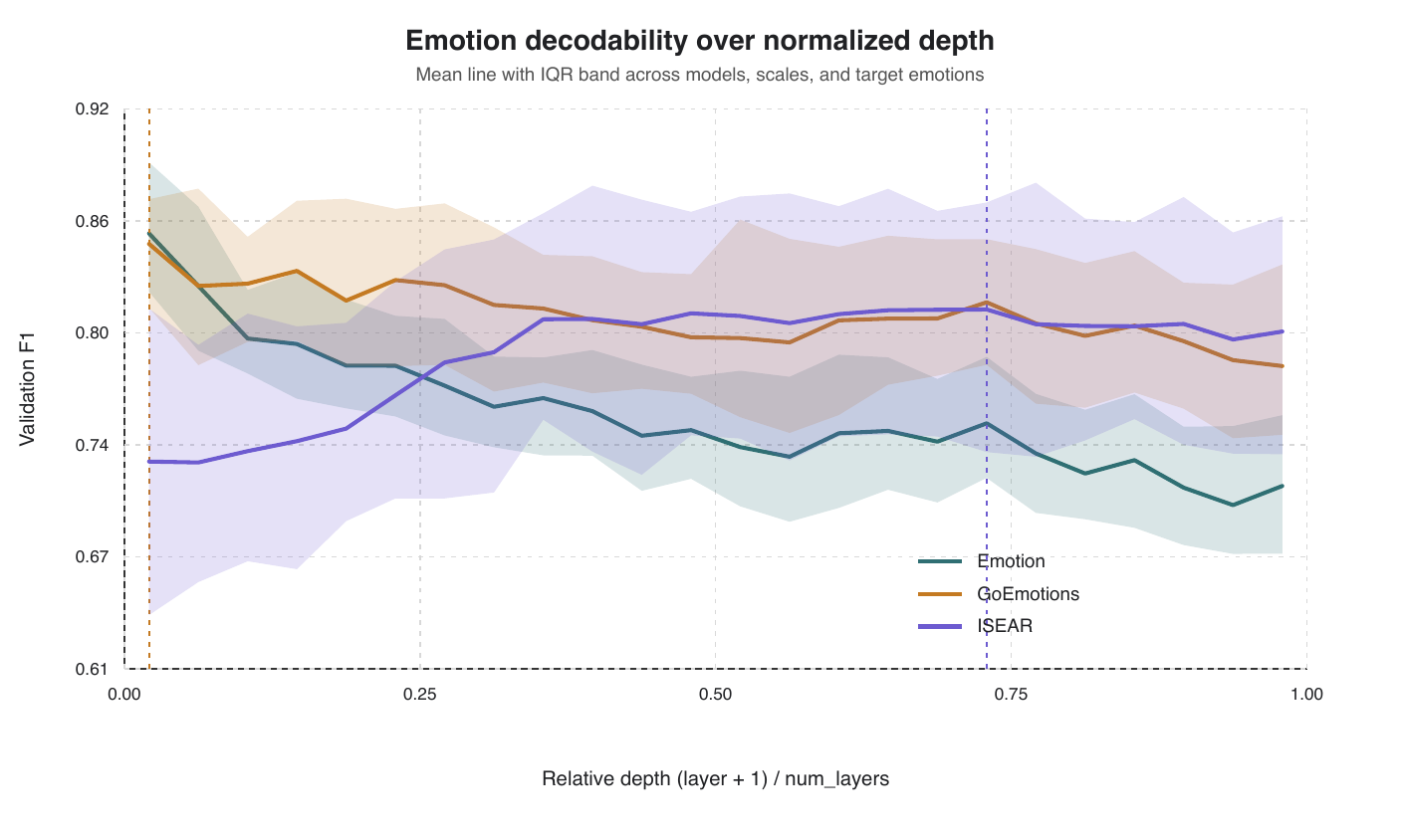}
  \caption{
  Dataset-level mean validation F1 over normalized depth, with IQR shading across models, scales, and target emotions. Emotion is early and sharp; GoEmotions is broader, more diffuse; ISEAR peaks substantially later.
  }
  \label{fig:layerwise-mean-iqr}
\end{figure*}

\noindent \textbf{Peak shape also differs across datasets.} Figure~\ref{fig:layerwise-mean-iqr} shows mean validation F1 with IQR shading over normalized depth. Emotion is early and sharply concentrated: layers within $0.01$ F1 of the peak span only $0.073$ normalized depth on average, or about $2.1$ raw layers. GoEmotions is much more diffuse ($0.316$ depth; $9.0$ layers), while ISEAR is later and moderately broad ($0.193$ depth; $5.9$ layers). Thus, datasets differ not only in peak location, but also in signal concentration.

\noindent \textbf{Emotions differ in difficulty and stability.} 
Joy is the most consistently decodable emotion, ranking first in all three datasets and in $19/24$ dataset--model contexts. Sadness is usually hardest, ranking last on Emotion and ISEAR and in $17/24$ dataset--model contexts, although it rises to second on GoEmotions. Fear is competitive on Emotion and ISEAR but weakest on GoEmotions, while anger is usually intermediate.

\noindent \textbf{OvR patterns hold under a four-class probe.} Because later interventions require target-emotion-specific bands, our main layer-wise analysis uses one-vs-rest probes. As a sanity check, we also train a four-class multinomial probe at each layer. Across 24 dataset--model pairs, four-class best depth strongly agrees with the mean OvR best depth ($r=0.85$), and $87.5\%$ of four-class best layers fall within the range of the four emotion-specific OvR best layers. Full results are in Appendix~\ref{app:layerwise}.

\noindent \textbf{Implications for intervention.}
These results motivate the intervention design in the next sections. First, because best depth varies systematically by dataset, layer selection must be performed on validation data within each dataset rather than using a universal raw layer index. Second, because the high-F1 region can be narrow or broad depending on the dataset, we test multiple band widths rather than a single best layer only. Third, if emotion-sensitive regions capture general emotional information, bands selected for one emotion or dataset should have measurable effects beyond their original selection context; we therefore test this hypothesis through transfer experiments.

\section{Causal Intervention on Emotion-Sensitive Layer Bands}
\label{sec:causal-intervention}

The layer-wise analysis (\S\ref{sec:layerwise}) shows that
emotion is linearly decodable at dataset-dependent depths. We next
ask whether these regions matter for downstream readout. We test this
with two complementary interventions: an \emph{online} intervention
that scales activations during the forward pass, and an \emph{offline}
intervention that scales the cached frozen-feature representation. The
online experiment provides the stronger causal evidence because the
perturbation can propagate through later layers; the offline experiment
checks whether the same bands matter in the static representation used
by the probe.

\subsection{Setup}
\label{sec:forward-setup}

For each model, dataset, target emotion, and band width
$w \in \{1,2,3,4\}$, we take the validation-selected emotion band from
the layer-wise analysis. We compare it against a same-width
\emph{transfer} band selected for the same emotion on the other
dataset, an \emph{unrelated} non-overlapping band, and an unmodified
baseline. During evaluation, a multiplicative hook scales the hidden
states in the band by $\alpha \in \{0.0,0.5,1.5\}$; $\alpha=0$
removes the band, $\alpha=0.5$ attenuates it, and $\alpha=1.5$
amplifies it. Both intervention experiments
evaluate Emotion and ISEAR.

All intervention claims use test-set predictions. We store per-example
predictions for each condition and compute paired correctness
contrasts within matched settings. Aggregate claims are summarized
over settings with bootstrap confidence intervals, sign tests,
Wilcoxon signed-rank tests, and BH-FDR correction within each
comparison family.

\begin{figure}[t]
\centering
\includegraphics[width=\linewidth]{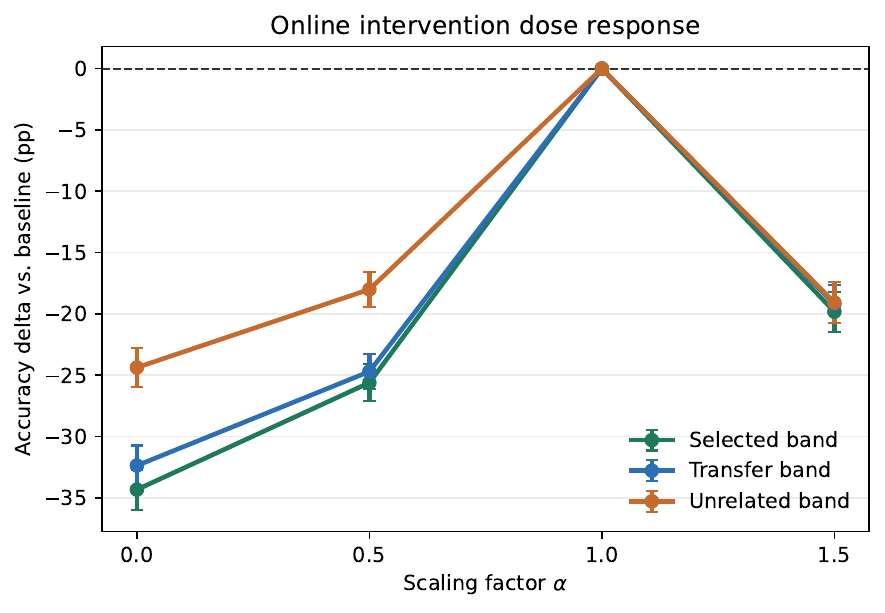}
\caption{Online intervention dose response. On average, selected and transferred
emotion bands are more disruptive than unrelated same-width bands,
especially when $\alpha < 1$. Values are test accuracy deltas relative
to the unmodified baseline. The $\alpha=1$ point denotes the unmodified
baseline and is shown only for reference.}
\label{fig:forward-alpha}
\end{figure}

\subsection{Online intervention: main effect and specificity}
\label{sec:forward-specificity}

Across all $768$ forward settings, scaling the emotion-selected band
causes a large accuracy drop relative to baseline
($-26.6$ percentage points, $q<0.001$). The unrelated-band control is
also disruptive ($-20.5$ points), showing that mid-forward
perturbations are not harmless. The key comparison is therefore
specificity: selected bands are more disruptive than unrelated
same-width bands by $-6.1$ points overall ($q<0.001$). At the strongest
intervention, $\alpha=0$, this specificity reaches $-8.2$, $-13.4$,
$-7.4$, and $-10.8$ points for widths $1$--$4$, respectively
(Table~\ref{tab:forward-specificity}). Thus, the layers identified by
validation probing are not merely decodable; perturbing them affects
the downstream readout more than perturbing arbitrary regions of the
same size.

\begin{table}[t]
\centering
\small
\begin{tabular}{lrrrr}
\hline
 & W1 & W2 & W3 & W4 \\
\hline
$\alpha=0.0$ & $-8.2^{*}$ & $-13.4^{***}$ & $-7.4$ & $-10.8^{**}$ \\
$\alpha=0.5$ & $-8.4^{*}$ & $-6.2$ & $-6.6$ & $-9.1$ \\
$\alpha=1.5$ & $-2.1$ & $-0.5$ & $-2.0$ & $+1.7$ \\
\hline
\end{tabular}
\caption{Forward specificity: selected-band minus unrelated-band
accuracy delta, in percentage points. Negative values mean selected
bands are more disruptive. Stars use BH-FDR-corrected sign tests:
$^{*}q<0.05$, $^{**}q<0.01$, $^{***}q<0.001$.}
\label{tab:forward-specificity}
\end{table}

\paragraph{Cross-dataset transfer.}
\label{sec:forward-transfer}
Transfer bands, selected for the same emotion on the other dataset,
produce nearly the same disruption as own-dataset bands. The
own-vs-transfer difference is small and not reliable overall
($-1.2$ points, n.s.), while transfer bands are still more disruptive
than unrelated bands ($-4.9$ points, $q<0.001$). A complementary $2\times2$ probe--band transfer control
(Appendix~\ref{app:probe_transfer}) compares target- and source-trained
LogReg probes under own and transferred bands. The own--transfer
difference remains approximately $-1.2$ points under either probe, and
the probe-by-band interaction is negligible ($-0.06$ points, n.s.),
indicating that the near-equivalence of own and transferred bands is
not an artifact of retaining the target-trained readout. This supports partial
cross-dataset sharing: an ISEAR-selected fear band, for example, can
remain causally relevant during an Emotion forward pass, although the
match is not exact.

\paragraph{Moderators.}
\label{sec:forward-moderators}
The specificity gap is strongest for Llama ($-7.8$ points) and Qwen
($-9.3$ points), both significant after correction, but not for
Granite ($+1.2$ points, n.s.). Dataset also matters. On Emotion,
selected bands are reliably more disruptive than unrelated bands
($-7.7$ points, $q<0.001$), with significant specificity at every
width. On ISEAR, the same contrast is smaller ($-4.5$ points) and less
stable. In contrast, transfer-vs-unrelated is strongest on ISEAR
($-12.2$ points), suggesting that deeper narrative-style cues may be
more compatible with transferred bands.

\subsection{Offline feature scaling}
\label{sec:fixed-check}
The offline intervention applies the same scaling operation after
feature extraction, directly to the cached representation read by the
probe. It therefore tests representational usefulness, not causal
computation inside the transformer. Offline scaling reproduces the
main disruption effect but with weaker specificity: own-band scaling
reduces accuracy by $-12.2$ points relative to baseline ($q<0.001$),
while unrelated and low-sensitivity bands also reduce accuracy by
$-11.3$ and $-10.8$ points. The selected-vs-unrelated contrast is only
$-0.8$ points and is not reliable. A complementary $2\times2$ probe--band transfer control shows that
the own--transfer difference remains similarly small under
source-trained and target-trained probes ($+0.6$ vs.\ $+0.7$ points),
with a negligible probe-by-band interaction ($-0.07$ points, n.s.;
Appendix~\ref{app:probe_transfer}). This contrast between online and
offline results suggests that the causal evidence comes primarily from
intervening during the forward computation, not merely from editing
the final feature vector.

\begin{figure}[t]
\centering
\includegraphics[width=\linewidth]{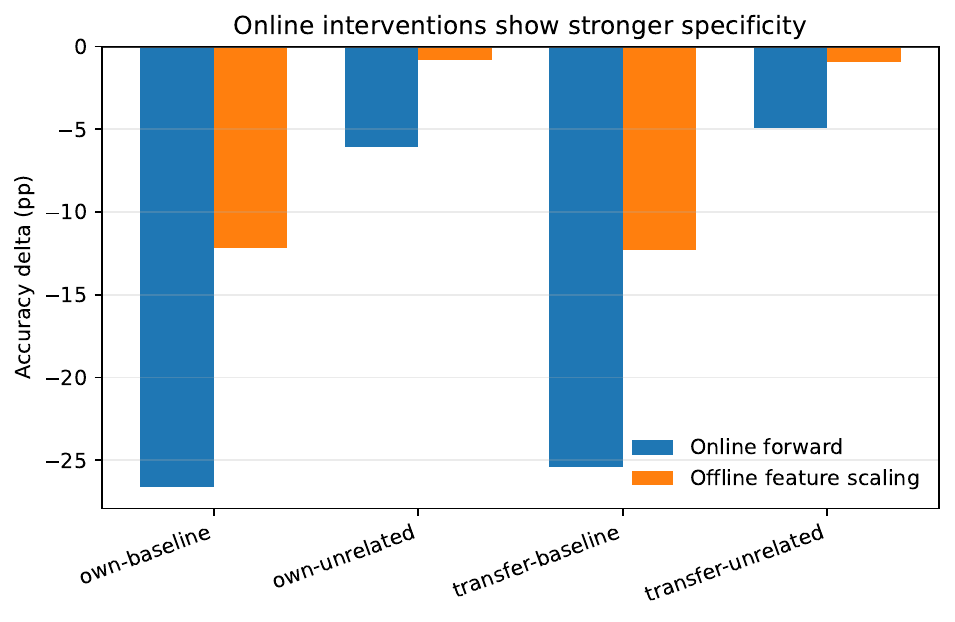}
\caption{Online vs.\ offline intervention effects. Both settings show
large disruption relative to baseline, but selected-vs-unrelated
specificity is much clearer online. In the offline setting, transfer
pools the available transfer conditions; own-vs-baseline and
own-vs-unrelated are directly comparable across settings.}
\label{fig:fixed-forward-comparison}
\end{figure}

\paragraph{Cross-emotion transfer.}
The offline setting additionally lets us test cross-emotion transfer:
scaling a band selected for a different emotion on the same dataset.
At $\alpha=0$, cross-emotion transfer is almost as disruptive as the
own-emotion band ($-15.2$ vs.\ $-15.9$ points). Fear is the most
vulnerable target ($-16.8$), joy the most robust ($-13.1$), and
sadness is the most disruptive source ($-16.5$ on average). These
patterns are exploratory, but they suggest that selected bands encode
shared affective structure rather than completely emotion-specific
modules.

% =========================================================
\section{Early Exit from Emotion-Sensitive Bands}
\label{sec:early-exit}

The intervention results show that emotion-selected bands are
functionally important for the probe-mediated readout. We next ask the
constructive question: can these bands support downstream
classification without running the full transformer?

\paragraph{Setup.}
\label{sec:early-exit-setup}
For each of the eight models and the two datasets evaluated here
(Emotion and ISEAR), we consider target-emotion bands of widths
$w\in\{1,2,3,4\}$. For each condition, we train a separate 4-class
multinomial logistic-regression head on the corresponding target-dataset
training representations. Features concatenate
\texttt{concat(mean,max,min)} pooled states across the layers in the
band. For selected, random, low-sensitivity, and cross-dataset transfer
bands, the forward pass stops after the final layer of the band. The
full-depth control runs the complete transformer and uses the same-width
band comprising the final $w$ layers. As above, all aggregate significance
tests use matched test-set predictions with BH-FDR correction.

\begin{figure}[t]
\centering
\includegraphics[width=\linewidth]{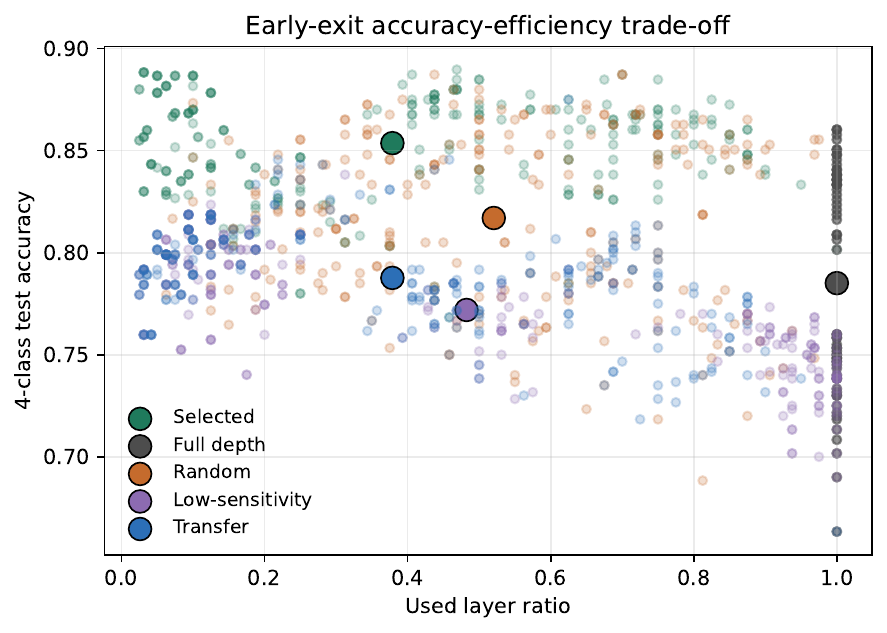}
\caption{Early-exit accuracy-efficiency trade-off. Each small point is
a model--dataset--emotion--width setting; large points show condition
means. Selected bands achieve higher accuracy while using fewer layers
than full-depth representations.}
\label{fig:early-exit-pareto}
\end{figure}

\paragraph{Selected bands outperform full depth and controls.}
\label{sec:early-exit-selected}
Selected early exits outperform full-depth representations by
$+6.9$ accuracy points overall ($q<0.001$). They also outperform
random bands ($+3.7$), low-sensitivity bands ($+8.2$), and transferred
bands ($+6.6$), all with $q<0.001$. These comparisons rule out the
trivial explanation that any intermediate representation works equally
well: the validation-selected emotion bands are consistently better
readout points for the lightweight classifier.

\paragraph{Bandwidth.}
\label{sec:early-exit-width}
Early-exit gains are not monotonic in bandwidth. Width $1$ gives the
largest selected-vs-full-depth gain ($+7.8$ points), width $3$ is the
weakest ($+5.9$), and width $4$ partially recovers ($+6.5$). This
matches the layer-wise picture: when emotion cues are concentrated,
a narrow band can capture the useful signal without adding off-peak
features; when cues are more distributed, wider bands add capacity but
with diminishing returns.

\paragraph{Early-exit transfer.}
\label{sec:early-exit-transfer}
Early exit exposes two notions of transfer. \emph{Layer transfer}
uses a band selected on the other dataset, while fitting a separate
classifier on target-dataset training examples for the representations
from that band. These transferred bands roughly match
full depth overall ($+0.3$ points, n.s.) but remain below
target-selected exits by $6.6$ points. The effect is asymmetric:
transfer exits help on Emotion relative to full depth ($+3.7$ points)
but hurt on ISEAR ($-3.2$ points). \emph{Source-train transfer} also
changes the classifier training data. In that setting, source-selected
bands slightly outperform target-selected bands under the
source-trained head ($+1.2$ points, $q<0.01$), suggesting that exit
quality depends both on where emotion is encoded and on which
representation the classifier was trained to read.

\paragraph{Frozen-encoder comparison.}
\label{sec:early-exit-baselines}
We also compare selected early-exit LLM representations with frozen
RoBERTa-base and DeBERTa-v3-base representations under the same
logistic-regression protocol. We keep all encoders frozen because our
goal is to evaluate where emotion-related information is already
available in pretrained representations, rather than how much performance
can be obtained after task-specific parameter updates. Fully fine-tuning
an encoder can reshape its internal representations, making it less
directly comparable to the layer-localization and early-exit setting
studied here.
Selected LLM exits outperform frozen RoBERTa by $+15.8$ points and
frozen DeBERTa by $+21.5$ points overall
(Table~\ref{tab:app-early-baselines}). This is a frozen-representation
comparison which shows that the selected
LLM bands provide a strong task representation under a lightweight
head.

\paragraph{Moderators.}
\label{sec:early-exit-moderators}
The dataset selection is the strongest moderator. Selected-vs-full-depth gain is
+11.3 points on Emotion but only +2.5 on ISEAR, consistent with
Emotion's shallow and concentrated layer-wise profile. Model family also
matters: Granite shows the largest gain (+8.8), followed by Llama
(+6.9) and Qwen (+5.5).

\section{Discussion}
\label{sec:discussion}

Overall, emotion-sensitive information is localizable enough for
intervention and early readout, but its layer-wise organization varies
with emotion expression and readout setting.

\noindent \textbf{Emotion depth reflects how affect is expressed.}
The clearest pattern is that the best emotion layers shift with dataset
source and style. This supports a more nuanced view than the claim that
affective information simply lives in early layers. Early layers are
sufficient for CARER-like short posts, where emotion is often marked by
lexical or conventional cues. In contrast, ISEAR narratives require
deeper representations, plausibly because the emotion must be inferred
from events, appraisals, and causal context. GoEmotions falls between
these cases, with a broader and less sharply localized profile. Thus,
the depth of emotion decodability should be treated as an empirical
property of the input distribution, not only as a property of the model.

\noindent \textbf{Intervention and readout reveal different notions of usefulness.}
The online intervention results show that probe-selected bands are not
only diagnostic: perturbing them disrupts downstream prediction more
than matched unrelated bands. However, the offline scaling results are
less specific, which suggests that the causal role of a band depends on
its participation in the forward computation, not only on the final
cached feature vector. Early-exit results add a constructive perspective:
some intermediate bands are better readout points than the final layer
under a lightweight classifier. Together, these findings separate three
questions that are often conflated: where emotion is decodable, where it
is causally involved, and where it is easiest to read out.

\noindent \textbf{Transfer is shared but readout-dependent.}
Transfer results show that emotion-sensitive regions are not strictly
dataset-specific or label-specific. Bands selected on one dataset can
remain disruptive on another, and different-emotion bands can affect the
same target. At the same time, early-exit transfer is weaker, indicating
that a useful causal region is not necessarily an optimal readout point
for a classifier trained in a different setting. This distinction helps
explain why transfer appears stronger in intervention than in early-exit
classification: the former tests whether a region matters during
computation, while the latter tests whether a trained classifier can use
that region effectively.

\noindent \textbf{Implications.}
These results suggest that affective information in LLMs is organized at
a band level rather than at a single universal layer. Future analyses of
emotion in LLMs should therefore report not only whether emotion is
decodable, but also the dataset source, model family, layer-selection
criterion, and readout mechanism under which the claim holds.

\section{Conclusion}
\label{sec:conclusion}

This paper examined whether emotion in decoder-only LLMs has a fixed
layer-wise location, or whether its organization varies with the kind of
text being processed. Our results support the latter. Emotion information
is not uniformly shallow, nor does it occupy a single universal layer
across models and datasets. Instead, the depth and breadth of
emotion-sensitive bands vary systematically with dataset style and model
family.

These bands are not merely correlational artifacts of a linear probe.
They can be causally perturbed through forward intervention and reused as
compact early-exit representations, while their cross-dataset and
cross-emotion transfer suggests shared affect-sensitive regions rather
than isolated label-specific modules.

Taken together, our findings recast emotion representation as a
depth-sensitive and context-dependent property of pretrained LLMs. This
view helps explain why prior layer-wise analyses can disagree across
datasets, and suggests that future work on affective behavior in LLMs
should treat where emotion is represented as part of the empirical
question, not as a fixed architectural fact.

\section*{Limitations}

Our analysis combines layer-wise probing, causal intervention, and
early-exit experiments on eight open-weight decoder-only LLMs and three
English emotion corpora under a fixed pooling and probe configuration.
Several aspects of this design constrain the generality of our
conclusions.

\paragraph{Dataset and language scope.}
We restrict our analysis to three English-language corpora
(Emotion/CARER, ISEAR, GoEmotions) and to the four-emotion intersection
of their original label sets: fear, joy, anger, and sadness. Other
emotions such as surprise, disgust, shame, guilt, and love are not
analyzed here. All three sources are English and largely Western, so
the dataset-depth ordering we observe may not generalize to other
languages, cultural contexts, or finer-grained emotion taxonomies.

\paragraph{Model scope.}
We probe eight open-weight decoder-only LLMs from three families
(Llama, Qwen, Granite) between 1B and 9B parameters. We do not evaluate
closed-weight models, base non-instruction-tuned checkpoints, or much
larger open models. The model-family and scale patterns should
therefore be read as descriptive findings within this open-weight
range, not as universal claims about all decoder-only LLMs.

\paragraph{Methodological choices.}
All main experiments keep model parameters frozen and use deterministic
pooling with a linear probe. This design
is intentional: our goal is to study where emotion-related information
is already available in pretrained representations, rather than how much
task-specific performance can be obtained after updating model
parameters. Fully fine-tuning an encoder or decoder could substantially
reshape internal representations, making it less directly comparable to
the layer-localization and early-exit setting studied here. A learnable
attention-pooling baseline can reach higher absolute F1 in pilot runs,
but it introduces additional training instability and capacity
differences, making representation comparisons harder to interpret. Our
interventions also operate at a coarse contiguous layer-band
granularity. Finer-grained approaches such as activation patching,
direction-level editing, nullspace projection, or neuron-level
interventions could localize affective computation more precisely.

\paragraph{Finding-level caveats.}
Forward specificity is significant for Llama and Qwen but not Granite.
Offline feature scaling reproduces the main disruption direction but
does not show reliable selected-vs-unrelated specificity, consistent
with our view that online intervention is the stronger causal
diagnostic. Cross-emotion transfer
(Table~\ref{tab:app-cross-emotion-transfer}) suggests shared affective
bands rather than strict per-emotion modules. The early-exit width
effect is non-monotonic, and our explanation in terms of selectivity
versus capacity remains exploratory. Source-train transfer is
consistent with a classifier--band alignment hypothesis, but does not
prove it.

\paragraph{Statistical and reproducibility caveats.}
We compute per-setting paired correctness contrasts from per-example
test predictions. Aggregate claims are summarized across settings using
bootstrap confidence intervals, sign tests, Wilcoxon signed-rank tests,
and BH-FDR correction within each comparison family. Per-setting
McNemar tests are available in the supplementary CSVs. We do not apply
more conservative family-wise correction across every comparison in the
paper. Finally, ISEAR is loaded through a community-uploaded HuggingFace
dataset, so exact ISEAR-specific reproducibility depends on that source
remaining unchanged.

\section*{Acknowledgments}
This work has received support from the investment programme "France 2030" as part of the IdEx programme (ANR-18-IDEX-0001) implemented by Université Paris Cité, under which the inIdEx project EFL is conducted.

\bibliography{custom}

\appendix

\section{Dataset Statistics and Corpus-Cue Diagnostics}
  \label{app:dataset_stats}

  This appendix reports the per-emotion sample distribution underlying
  the aggregate dataset sizes in Table~\ref{tab:dataset_sizes}. 

  \begin{table*}[h]
  \centering
  \small
  \begin{tabular}{llrrrrr}
  \hline
  Dataset & Split & Fear & Joy & Anger & Sadness & Total \\
  \hline
  \multirow{3}{*}{Emotion (CARER)}
         & Train & 1{,}200 & 1{,}200 & 1{,}200 & 1{,}200 & 4{,}800 \\
  \cline{2-7}
         & Val   & 150     & 150     & 150     & 150     & 600 \\
  \cline{2-7}
         & Test  & 150     & 150     & 150     & 150     & 600 \\
  \hline
  \multirow{3}{*}{ISEAR}
         & Train & 823     & 816     & 831     & 764     & 3{,}234 \\
  \cline{2-7}
         & Val   & 102     & 102     & 103     & 95      & 402 \\
  \cline{2-7}
         & Test  & 104     & 102     & 105     & 97      & 408 \\
  \hline
  \multirow{3}{*}{GoEmotions}
         & Train & 321     & 604     & 644     & 578     & 2{,}147 \\
  \cline{2-7}
         & Val   & 68      & 129     & 138     & 124     & 459 \\
  \cline{2-7}
         & Test  & 70      & 130     & 138     & 125     & 463 \\
  \hline
  \end{tabular}
  \caption{Per-emotion sample counts in the four-emotion splits of
  each dataset.}
  \label{tab:per_emotion_stats}
  \end{table*}

  \paragraph{Corpus-cue diagnostics.}
The three corpora share the same four target labels but differ in
source, elicitation procedure, and surface-cue profile. To make cue
rates comparable despite differences in text length, we compute these
diagnostics on the label-by-length-bin matched training subsets. Each dataset contributes
1{,}909 matched training examples. Table~\ref{tab:matched_cue_profiles}
reports percentages macro-averaged across the four target emotions.

\paragraph{Operational definitions.}
Texts are lowercased and tokenized using an alphabetic word pattern
that preserves internal apostrophes. An \emph{exact target-label word}
is present when the gold-label token itself (e.g., \textit{fear} for a
fear example) occurs in the text. The \emph{emotion-lexicon} indicator
records whether at least one token belongs to the union of our
hand-specified fear, joy, anger, and sadness lexicons. A
\emph{first-person feel-form frame} requires one of
\{\textit{feel, feels, feeling, felt}\} to occur within three tokens
after a first-person subject such as \textit{I} or \textit{we}.
\emph{Second-person reference} records the presence of a second-person
pronoun or common informal variant. The \emph{temporal-marker}
indicator uses a predefined list of temporal words and phrases, such
as \textit{when}, \textit{after}, \textit{during}, and
\textit{at that moment}. The \emph{past-event proxy} records either a
predefined irregular-past form (e.g., \textit{was}, \textit{had}, or
\textit{felt}) or a token longer than three characters ending in
\texttt{-ed}. Full word and phrase lists are provided with the released
diagnostic code.

\begin{table*}[t]
\centering
\small
\begin{tabular}{lrrr}
\hline
Cue (\% of matched texts) & Emotion & GoEmotions & ISEAR \\
\hline
Exact target-label word             & 0.7  & 1.7  & 4.8  \\
Any heuristic emotion-lexicon cue   & 31.0 & 41.1 & 22.0 \\
First-person feel/felt frame        & 75.8 & 2.4  & 7.7  \\
Second-person reference             & 6.0  & 23.5 & 1.3  \\
Temporal marker                     & 14.9 & 12.8 & 62.4 \\
Past-event proxy                    & 56.9 & 43.4 & 82.9 \\
\hline
\end{tabular}
\caption{Corpus-cue diagnostics on label-by-length-bin matched training
subsets, macro-averaged across the four target emotions. These
string-pattern diagnostics are descriptive proxies rather than
linguistic annotations.}
\label{tab:matched_cue_profiles}
\end{table*}

The cue profiles do not form a single monotonic continuum. Exact-label
mentions increase from Emotion to ISEAR, whereas emotion-lexicon
coverage peaks for GoEmotions and first-person feel/felt frames are most
frequent in Emotion. We therefore treat dataset source as a
multidimensional property of the input distribution rather than as a
scalar measure of linguistic complexity.

\section{Pooling and Probe Selection Pilot}
  \label{app:pooling_probe_pilot}

  This appendix details the two-step greedy selection protocol used
  to fix the pooling function and probe classifier. All reported numbers are validation
  F1 at the best-performing layer per (emotion, pooling, probe) cell.

  \paragraph{Step 1: Pooling selection.}
  We fix the probe to a linear SVM on Llama-3.2-3B-Instruct and scan
  all layers of the Emotion validation split for each of the four
  target emotions, comparing last-token, mean, and
  \texttt{concat(mean, max, min)} pooling
  (Table~\ref{tab:pooling_pilot}). \texttt{concat} is the strongest
  non-parametric option on all four emotions, outperforming
  last-token by 24 to 35 F1 points and mean by 2 to 8 F1 points.
  
  \begin{table}[h]
  \centering
  \small
  \begin{tabular}{lrrrrr}
  \hline
  Pooling     & Fear  & Joy   & Anger & Sadness & Mean \\
  \hline
  last-token  & 0.554 & 0.628 & 0.551 & 0.494 & 0.557 \\
  \hline
  mean        & 0.829 & 0.876 & 0.794 & 0.743 & 0.811 \\
  \hline
  concat      & 0.908 & 0.880 & 0.836 & 0.766 & 0.848 \\
  \hline
  \end{tabular}
  \caption{Step-1 pooling pilot on Llama-3.2-3B-Instruct, Emotion
  validation split, linear SVM probe. Best-layer F1 per emotion.}
  \label{tab:pooling_pilot}
  \end{table}

  \paragraph{Max/min ablation.}
  To verify that the \texttt{concat} advantage is not driven by the
  max or min component alone, we additionally compare \texttt{concat}
  against max and min pooling on fear and joy with the same Llama-3B
  linear SVM probe (Table~\ref{tab:maxmin_ablation}). \texttt{concat}
  remains the strongest option in both cases.

  \begin{table}[h]
  \centering
  \small
  \begin{tabular}{lrr}
  \hline
  Pooling & Fear & Joy \\
  \hline
  concat & 0.908 & 0.880 \\
  \hline
  max    & 0.878 & 0.818 \\
  \hline
  min    & 0.843 & 0.836 \\
  \hline
  \end{tabular}
  \caption{Max/min ablation on Llama-3.2-3B-Instruct, Emotion
  validation split, linear SVM probe.}
  \label{tab:maxmin_ablation}
  \end{table}

  \paragraph{Cross-family pooling sanity check.}
  We replicate the pooling comparison on Qwen-3.5-4B for fear (one
  emotion only). The ranking is consistent with Llama-3B:
  \texttt{concat} (F1 $0.822$) is strongest, followed by mean
  ($0.795$) and last-token ($0.536$).

  \paragraph{Step 2: Probe selection.}
  With \texttt{concat(mean, max, min)} pooling fixed, we compare
  three lightweight probes on the same Llama-3B four-emotion
  validation setup: linear SVM, logistic regression, and a
  one-hidden-layer MLP (Table~\ref{tab:probe_pilot}). Logistic
  regression is the strongest probe on fear and joy and competitive
  on anger and sadness (within $0.01$ F1 of the best); we adopt it
  as the fixed analysis probe because it is also simpler and more
  stable than the MLP alternative.

  \begin{table}[h]
  \centering
  \small
  \begin{tabular}{lrrrrr}
  \hline
  Probe       & Fear  & Joy   & Anger & Sadness & Mean \\
  \hline
  linear SVM  & 0.885 & 0.858 & 0.833 & 0.777 & 0.838 \\
  \hline
  logreg      & 0.903 & 0.900 & 0.837 & 0.800 & 0.860 \\
  \hline
  MLP         & 0.893 & 0.858 & 0.845 & 0.812 & 0.852 \\
  \hline
  \end{tabular}
  \caption{Step-2 probe pilot on Llama-3.2-3B-Instruct, Emotion
  validation split, \texttt{concat(mean, max, min)} pooling.
  Best-layer F1 per emotion.}
  \label{tab:probe_pilot}
  \end{table}

  \paragraph{Cross-family probe sanity check.}
  We replicate the probe comparison on Qwen-3.5-4B for fear: linear
  SVM $0.843$, logreg $0.847$, MLP $0.832$. Logistic regression is
  again the strongest probe.

  \paragraph{Verification: learned attention pooling.}
  As a verification of our restriction to non-parametric pooling, we evaluate a single-head learnable
  attention pooling layer followed by a linear classification head,
  trained end-to-end with cross-entropy on the
  Llama-3.2-3B-Instruct Emotion validation split. We use the same
  random seed ($42$) as our main experiments, a fixed learning rate
  of $\eta=0.003$, and sweep over weight decay
  $\{0, 10^{-4}, 10^{-3}\}$ at each of six representative layers
  ($\{0, 3, 4, 20, 23, 26\}$), yielding $72$ runs in total.

  Table~\ref{tab:attention_matrix} reports F1 per (layer, emotion)
  averaged across the three weight decay values. At layer 0,
  attention pooling reaches strong F1 on all four emotions, but
  training collapses in $10$ of the $24$ (layer, emotion) cells at
  this seed: joy fails from layer 3 onward, fear from layer 20, and
  most emotions are reduced to $F1 = 0$ by layer 23. The mean F1
  across the $24$ cells is $0.51$, well below the concat baseline
  mean of $0.85$.

   \begin{table}[h]
  \centering
  \small
  \begin{tabular}{rrrrr}
  \hline
  Layer & Fear  & Joy   & Anger & Sadness \\
  \hline
  0  & 0.95 & 0.96 & 0.93 & 0.93 \\
  \hline
  3  & 0.94 & 0.00 & 0.94 & 0.96 \\
  \hline
  4  & 0.59 & 0.00 & 0.94 & 0.94 \\
  \hline
  20 & 0.00 & 0.00 & 0.94 & 0.92 \\
  \hline
  23 & 0.00 & 0.00 & 0.92 & 0.00 \\
  \hline
  26 & 0.00 & 0.00 & 0.40 & 0.00 \\
  \hline
  \end{tabular}
  \caption{Per-(layer, emotion) F1 of attention pooling at
  $\eta=0.003$, seed 42, averaged across three weight decay values.
  Zero entries indicate training collapse.}
  \label{tab:attention_matrix}
  \end{table}

  For completeness, when we cherry-pick the best (layer, weight
  decay) per emotion at this seed, the best-case mean F1 reaches
  $0.96$ (fear $0.957$, joy $0.963$, anger $0.946$, sadness
  $0.961$), exceeding concat on each emotion. Achieving this
  best-case in practice would require either layer-by-layer
  collapse screening or multi-seed retries to find a configuration
  that converges, neither of which is feasible at the scale of our
  main experiments ($8$ models $\times$ $3$ datasets $\times$ all
  transformer layers). In contrast, \texttt{concat(mean, max, min)}
  pooling is fully deterministic and applies uniformly without
  retuning. This per-layer instability also reflects the
  methodological cost of introducing learned parameters in the
  pooling stage: per-layer F1 measures not only the representation
  but also whether the attention parameters successfully trained at
  that layer.

\section{Additional Layer-Wise Analyses}
\label{app:layerwise}

This appendix provides supplementary analyses for Section~\ref{sec:layerwise}. All results are computed from validation-only layer-wise probing outputs. The main paper reports the condensed dataset- and family-level patterns; here we include the full one-vs-rest profiles, the four-class consistency check, cross-emotion correlations, and additional descriptive summaries.

\begin{figure*}[t]
  \centering
  % \includesvg[width=\textwidth,inkscapelatex=false]{official/layerwise_figures/layerwise_normalized_depth}
  \includegraphics[width=\textwidth]{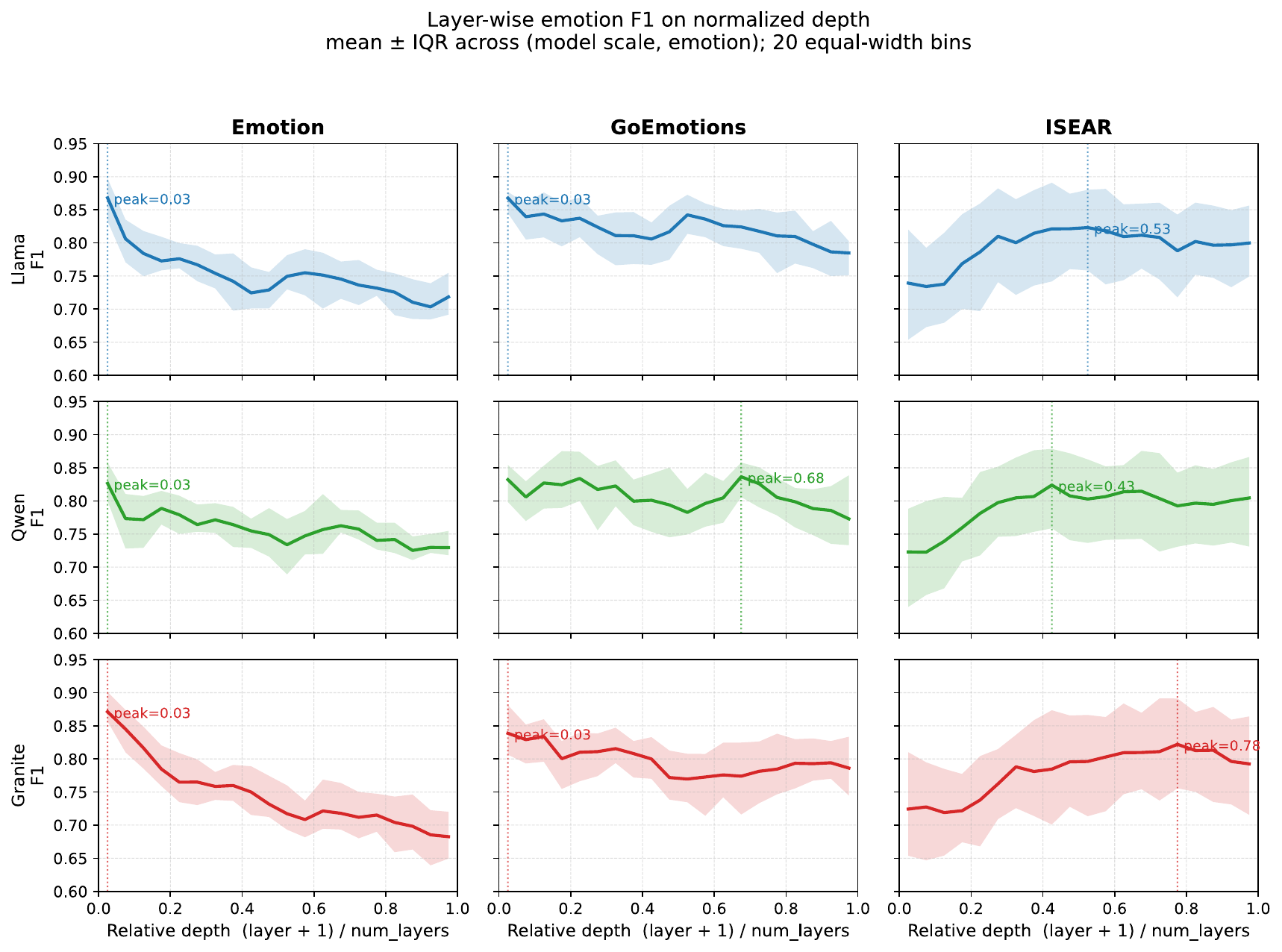}
  \caption{
  Full family-by-dataset normalized-depth profiles. Each panel aggregates validation F1 over model scales and target emotions within one model family and dataset.
  }
  \label{fig:app-layerwise-full}
\end{figure*}

\paragraph{Full normalized-depth profiles.}
Figure~\ref{fig:app-layerwise-full} shows the full $3\times3$ family-by-dataset normalized-depth view. This is the expanded version of Figure~\ref{fig:layerwise-family-dataset}. The same dataset-level ordering remains visible within model families: Emotion peaks early, GoEmotions is broader and intermediate, and ISEAR peaks later.

\begin{figure*}[t]
  \centering
  % \includesvg[width=\textwidth,inkscapelatex=false]{official/layerwise_figures_normalized/emotion_layerwise_panels_normalized}
  \includegraphics[width=\textwidth]{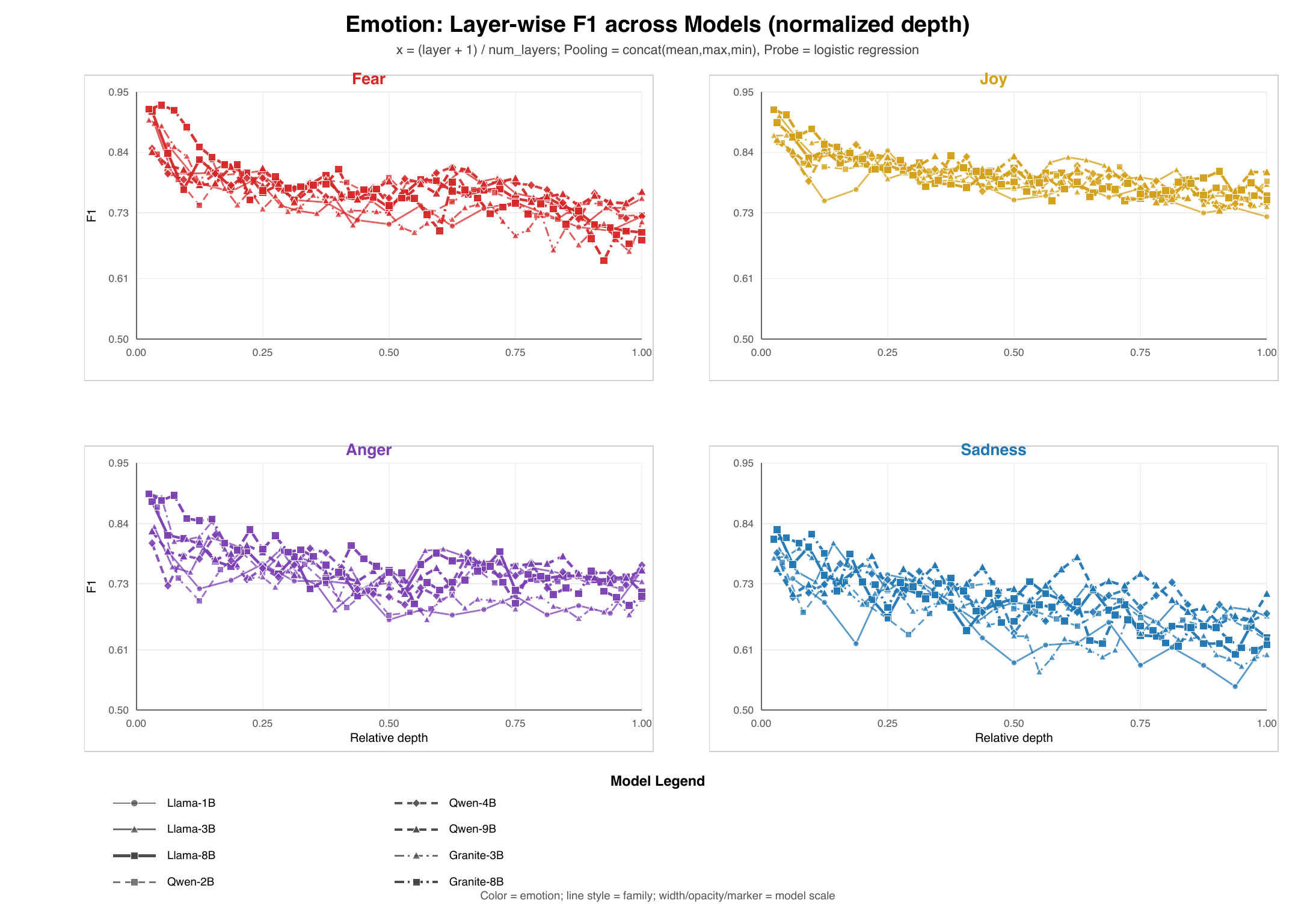}
  \caption{
  Emotion dataset: emotion-specific layer-wise validation F1 over normalized depth.
  }
  \label{fig:app-layerwise-emotion}
\end{figure*}

\begin{figure*}[t]
  \centering
  \includegraphics[width=\textwidth]{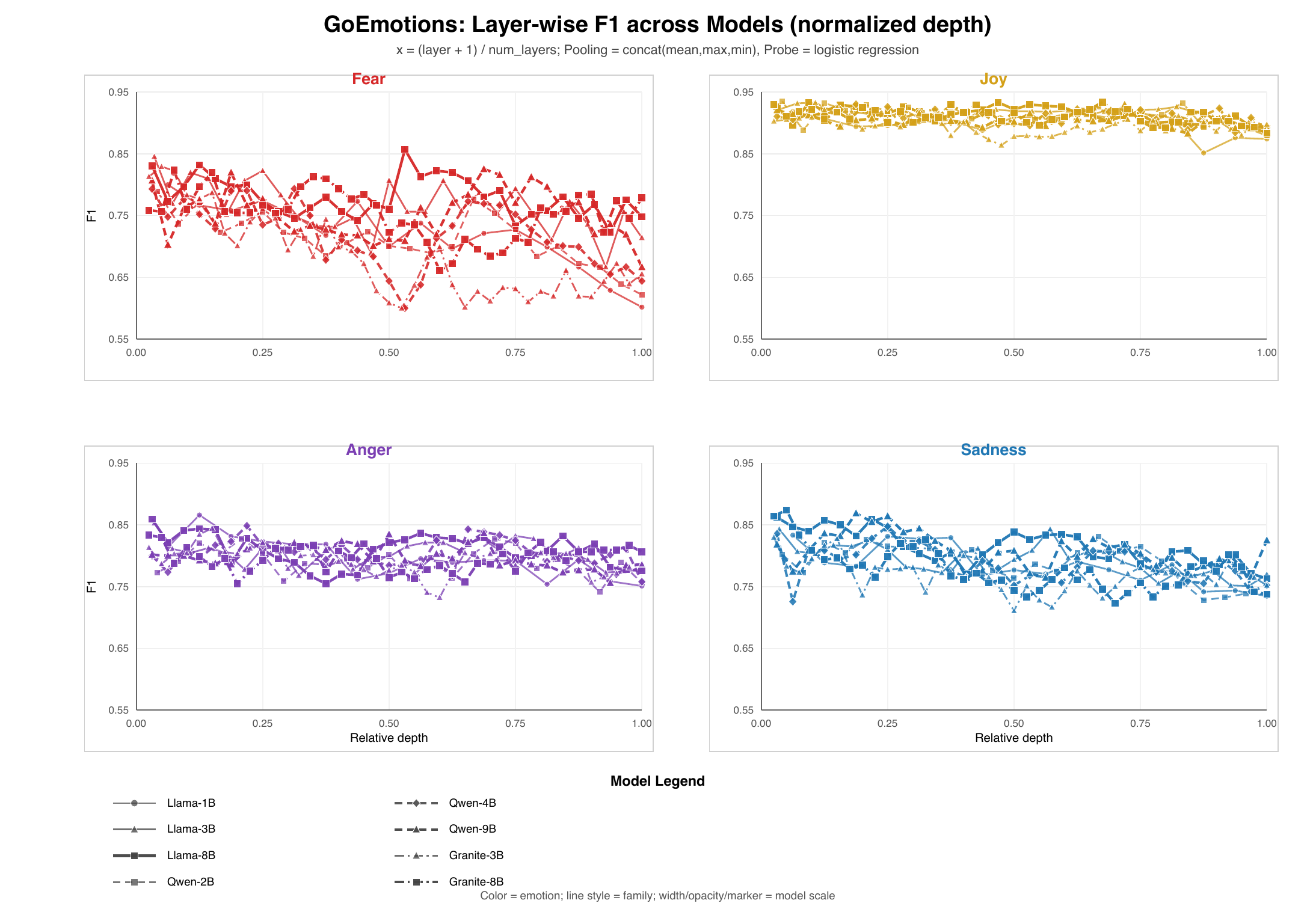}
  \caption{
  GoEmotions: emotion-specific layer-wise validation F1 over normalized depth.
  }
  \label{fig:app-layerwise-goemotions}
\end{figure*}

\begin{figure*}[t]
  \centering
  \includegraphics[width=\textwidth]{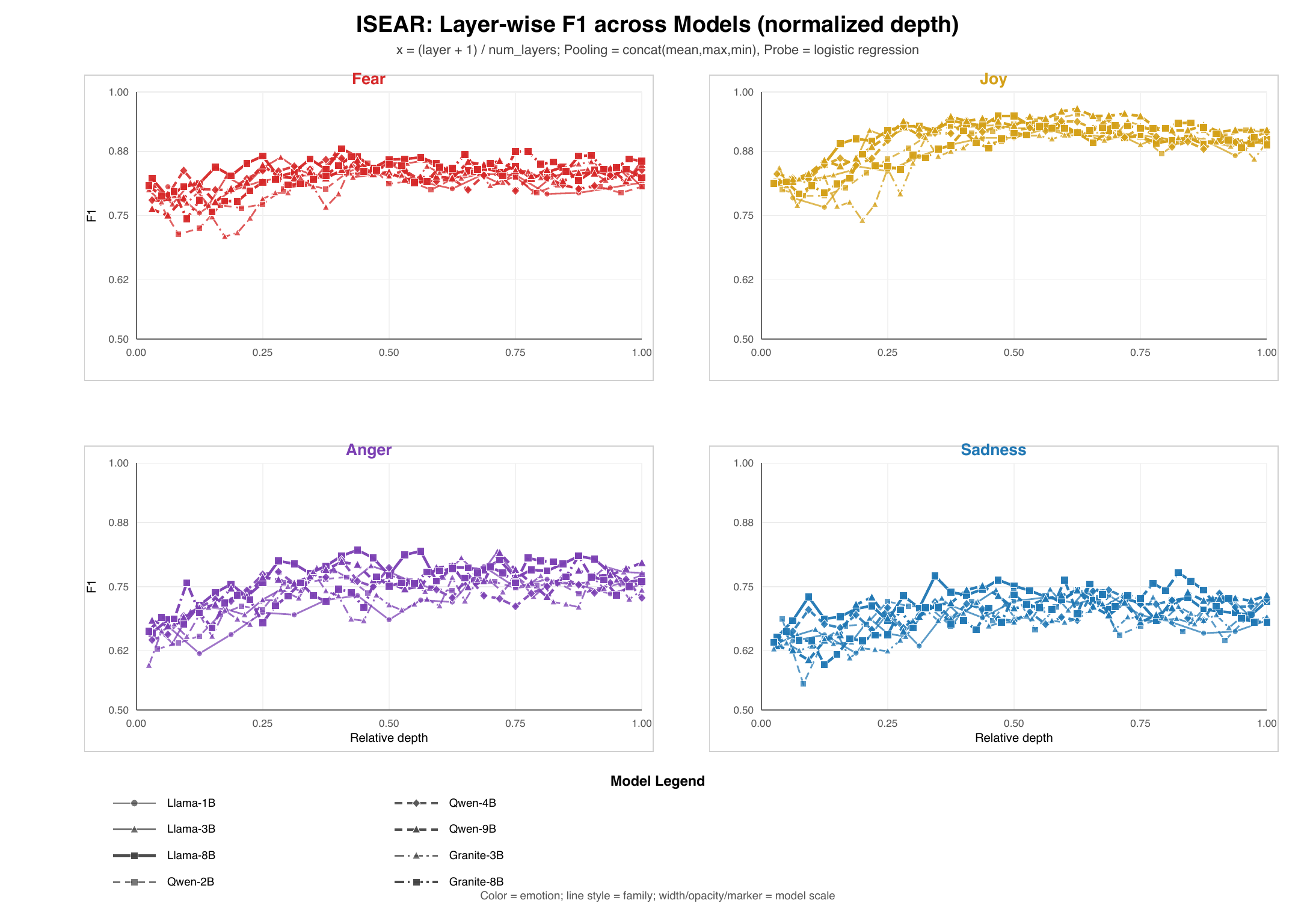}
  \caption{
  ISEAR: emotion-specific layer-wise validation F1 over normalized depth.
  }
  \label{fig:app-layerwise-isear}
\end{figure*}

\paragraph{Emotion-specific layer-wise profiles.}
Figures~\ref{fig:app-layerwise-emotion}, \ref{fig:app-layerwise-goemotions}, and \ref{fig:app-layerwise-isear} show the normalized-depth profiles separately for each target emotion. These figures provide the detailed curves underlying the dataset-level summaries in the main text.

\begin{figure*}[t]
  \centering
  \includegraphics[width=.72\linewidth]{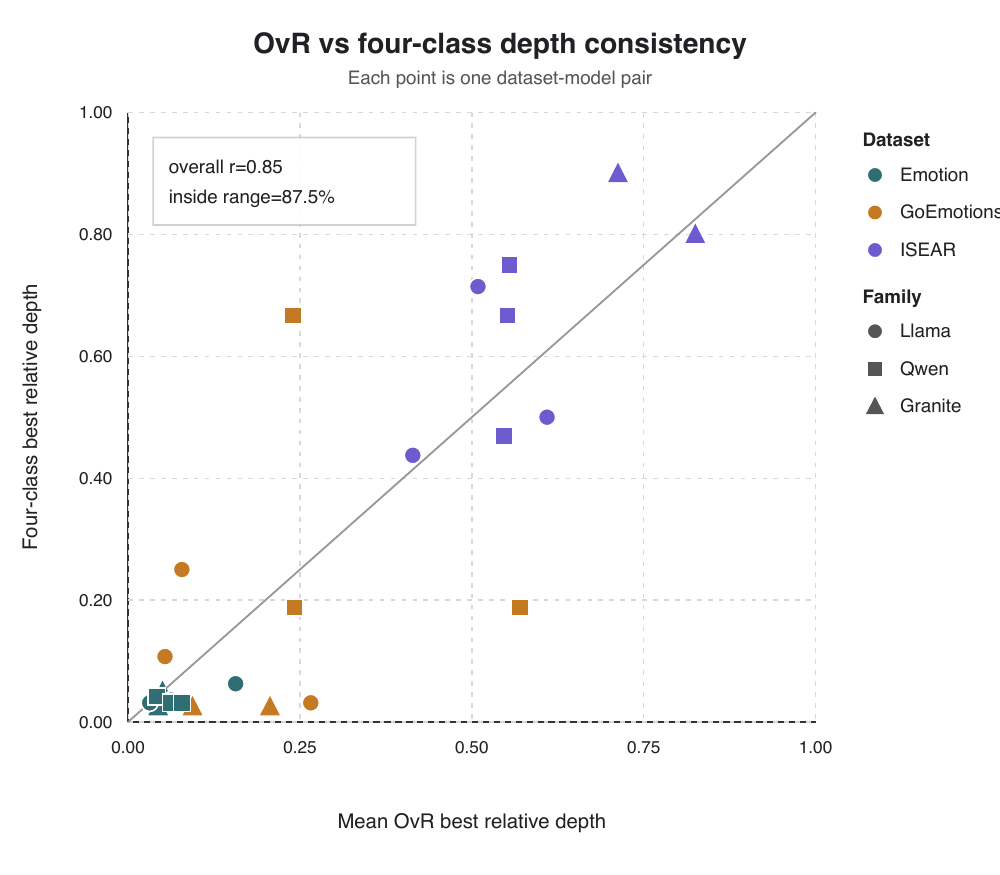}
  \caption{
  Consistency between one-vs-rest and four-class layer-wise probes. Each point is a dataset--model pair.
  }
  \label{fig:app-ovr-multiclass}
\end{figure*}

\paragraph{OvR versus four-class consistency.}
Figure~\ref{fig:app-ovr-multiclass} compares the best relative depth from the four-class multinomial probe against the mean best relative depth from the four one-vs-rest probes. Across all 24 dataset--model pairs, the correlation is $r=0.85$, and $87.5\%$ of four-class best layers fall within the range spanned by the four one-vs-rest best layers. This supports using one-vs-rest probes for target-specific band selection.

\section{Forward Intervention Significance Details}
\label{app:forward_significance}

The forward master CSV contains $768$ matched test settings
($8$ models $\times$ $2$ datasets $\times$ $4$ emotions $\times$
$4$ widths $\times$ $3$ alpha values). For each comparison, we compute
paired correctness contrasts from per-example predictions, then
summarize across settings with bootstrap confidence intervals, sign
tests, Wilcoxon signed-rank tests, and BH-FDR correction.

\begin{table}[h]
\centering
\small
\begin{tabular}{lrr}
\hline
Comparison & $n$ & Mean $\Delta$acc \\
\hline
Own vs.\ baseline & $768$ & $-26.6^{***}$ \\
Transfer vs.\ baseline & $768$ & $-25.4^{***}$ \\
Unrelated vs.\ baseline & $768$ & $-20.5^{***}$ \\
Own vs.\ transfer & $768$ & $-1.2$ \\
Own vs.\ unrelated & $768$ & $-6.1^{***}$ \\
Transfer vs.\ unrelated & $768$ & $-4.9^{***}$ \\
\hline
\end{tabular}
\caption{Forward intervention aggregate effects in percentage points.
Stars indicate BH-FDR-corrected significance.}
\label{tab:app-forward-main}
\end{table}

\section{Offline Feature-Scaling Significance Details}
\label{app:fixed_significance}

The offline master CSV mirrors the forward setup but applies scaling
to cached representations. It contains own/control settings plus
cross-dataset and cross-emotion transfer settings. Because the
transformer is not re-run after the edit, this analysis supports
representational usefulness but is weaker causal evidence than the
online intervention.

\begin{table}[h]
\centering
\small
\begin{tabular}{lrr}
\hline
Comparison & $n$ & Mean $\Delta$acc \\
\hline
Own vs.\ baseline & $768$ & $-12.2^{***}$ \\
Transfer vs.\ baseline & $3072$ & $-12.3^{***}$ \\
Unrelated vs.\ baseline & $768$ & $-11.3^{***}$ \\
Low-sensitivity vs.\ baseline & $768$ & $-10.8^{***}$ \\
Own vs.\ unrelated & $768$ & $-0.8$ \\
Own vs.\ low-sensitivity & $768$ & $-1.3$ \\
Transfer vs.\ own & $3072$ & $-0.1$ \\
Transfer vs.\ unrelated & $3072$ & $-1.0$ \\
\hline
\end{tabular}
\caption{Offline feature-scaling aggregate effects in percentage
points. Main disruption effects are reliable, but selected-band
specificity is weak.}
\label{tab:app-fixed-main}
\end{table}

\begin{table}[h]
\centering
\small
\setlength{\tabcolsep}{3pt}
\begin{tabular}{lrrrrr}
\hline
Target / Source & Anger & Fear & Joy & Sad. & Mean \\
\hline
Anger & -- & $-14.9$ & $-14.6$ & $-16.6$ & $-15.4$ \\
Fear & $-17.0$ & -- & $-16.7$ & $-16.8$ & $-16.8$ \\
Joy & $-11.2$ & $-12.0$ & -- & $-16.1$ & $-13.1$ \\
Sad. & $-14.3$ & $-16.4$ & $-15.7$ & -- & $-15.5$ \\
\hline
Mean & $-14.2$ & $-14.5$ & $-15.7$ & $-16.5$ & \\
\hline
\end{tabular}
\caption{Offline cross-emotion transfer at $\alpha=0$, in accuracy
points. Row is the target probe; column is the emotion used to select
the perturbed source band.}
\label{tab:app-cross-emotion-transfer}
\end{table}

\section{Early Exit Significance Details}
\label{app:early_exit_significance}

The early-exit master CSV combines target-trained controls,
cross-dataset layer-transfer runs, and source-trained transfer runs.
All reported tests use matched test-set predictions and BH-FDR
correction within comparison families.

\begin{table}[h]
\centering
\small
\begin{tabular}{lrr}
\hline
Comparison & $n$ & Mean $\Delta$acc \\
\hline
Selected vs.\ full depth & $256$ & $+6.9^{***}$ \\
Selected vs.\ random & $256$ & $+3.7^{***}$ \\
Selected vs.\ low-sensitivity & $256$ & $+8.2^{***}$ \\
Selected vs.\ transfer & $256$ & $+6.6^{***}$ \\
Transfer vs.\ full depth & $256$ & $+0.3$ \\
\hline
\end{tabular}
\caption{Early-exit aggregate effects in percentage points. Positive
values mean the first condition is more accurate.}
\label{tab:app-early-main}
\end{table}

\begin{table}[h]
\centering
\small
\begin{tabular}{lrr}
\hline
Stratum & $n$ & Selected vs.\ full depth \\
\hline
Emotion & $128$ & $+11.3^{***}$ \\
ISEAR & $128$ & $+2.5^{***}$ \\
Granite & $64$ & $+8.8^{***}$ \\
Llama & $96$ & $+6.9^{***}$ \\
Qwen & $96$ & $+5.5^{***}$ \\
\hline
\end{tabular}
\caption{Early-exit selected-vs-full-depth gain by dataset and model
family, in accuracy points.}
\label{tab:app-early-moderators}
\end{table}

\begin{figure*}[t]
\centering
\includegraphics[width=\linewidth]{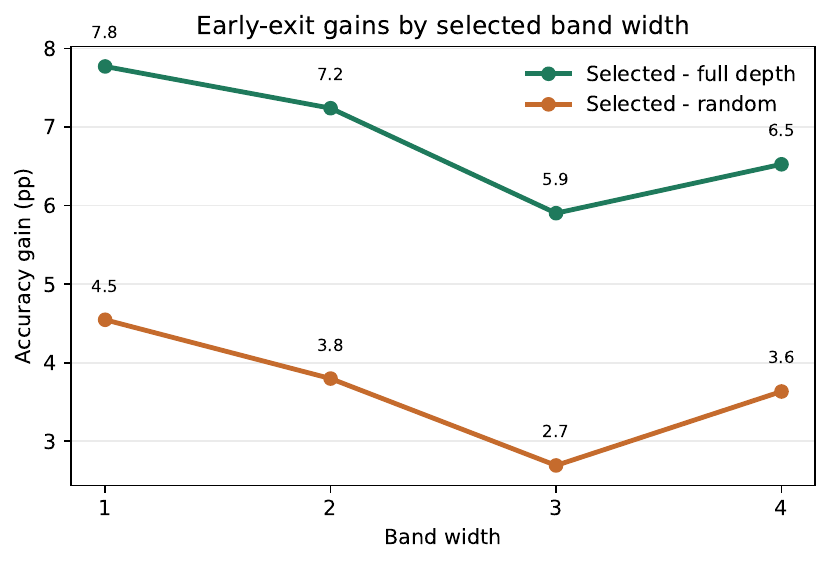}
\caption{Early-exit gains by band width. Selected bands outperform
both full-depth and random controls at every width, with the largest
gain at width $1$.}
\label{fig:early-exit-width}
\end{figure*}

\begin{table}[h]
\centering
\small
\begin{tabular}{lrr}
\hline
Width & Selected vs.\ full depth & Selected vs.\ random \\
\hline
1 & $+7.8^{***}$ & $+4.5^{***}$ \\
2 & $+7.2^{***}$ & $+3.8^{***}$ \\
3 & $+5.9^{***}$ & $+2.7^{***}$ \\
4 & $+6.5^{***}$ & $+3.6^{***}$ \\
\hline
\end{tabular}
\caption{Early-exit gains by band width, in accuracy points.}
\label{tab:app-early-width}
\end{table}

\begin{table}[h]
\centering
\small
\begin{tabular}{lrr}
\hline
Baseline & $n$ & Selected LLM exit gain \\
\hline
Frozen RoBERTa-base & $256$ & $+15.8^{***}$ \\
Frozen DeBERTa-v3-base & $256$ & $+21.5^{***}$ \\
\hline
\end{tabular}
\caption{Frozen encoder baseline comparison under the same
logistic-regression readout.}
\label{tab:app-early-baselines}
\end{table}

\section{Joint Probe and Layer-Band Transfer}
\label{app:probe_transfer}

The main transfer experiments change the selected layer band while
retaining the probe trained on the target dataset. To determine whether
the resulting transfer effects depend on this target-trained readout,
we conduct a $2\times2$ probe--band transfer control. We independently
vary:

\begin{enumerate}[noitemsep,topsep=0pt]
    \item the probe-training dataset: target-trained versus
    source-trained logistic regression; and
    \item the intervened band: the target dataset's own band versus the
    same-emotion band selected on the source dataset.
\end{enumerate}

Both probes are evaluated on the target test set. The source-trained
probe, including its \texttt{StandardScaler}, is fitted exclusively on
the paired source dataset and applied to the target examples without
refitting. We run the same $2\times2$ design for both the online
forward intervention and the offline cached-feature intervention,
covering eight models, both transfer directions
(Emotion$\leftrightarrow$ISEAR), four emotions, four band widths, and
three intervention strengths, for $768$ matched settings per
experiment.

For probe $p\in\{\mathrm{source},\mathrm{target}\}$, we define its
within-probe band preference as
\[
D_p =
\operatorname{Acc}(p,\mathrm{own\ band})
-
\operatorname{Acc}(p,\mathrm{transferred\ band}),
\]
and the probe-by-band interaction as
\[
I = D_{\mathrm{source}}-D_{\mathrm{target}}.
\]
A nonzero interaction would indicate that the relative effect of the
two bands depends on which dataset trained the readout.

\begin{table*}[t]
\centering
\small
\begin{tabular}{lllrrr}
\hline
Intervention & Metric & Probe &
Own$-$Baseline & Transfer$-$Baseline & Own$-$Transfer \\
\hline
Online
& Accuracy & Source & $-16.31$ & $-15.09$ & $-1.22$ \\
& Accuracy & Target & $-26.59$ & $-25.42$ & $-1.16$ \\
& F1       & Source & $-21.68$ & $-22.07$ & $+0.38$ \\
& F1       & Target & $-35.44$ & $-36.37$ & $+0.92$ \\
\hline
Offline
& Accuracy & Source & $-6.79$  & $-7.43$  & $+0.64$ \\
& Accuracy & Target & $-12.12$ & $-12.82$ & $+0.71$ \\
& F1       & Source & $-10.17$ & $-9.76$  & $-0.40$ \\
& F1       & Target & $-14.34$ & $-15.44$ & $+1.11$ \\
\hline
\end{tabular}
\caption{Joint probe--band transfer results, averaged over the
$768$ matched settings. Values are percentage-point differences.
Within both source- and target-trained probes, own and transferred
bands produce similar effects, although absolute disruption is larger
under the target-trained probe.}
\label{tab:probe_band_transfer}
\end{table*}

\begin{table}[t]
\centering
\small
\begin{tabular}{lrrr}
\hline
Intervention / Metric &
Interaction & Setting $q$ & Model-cluster $q$ \\
\hline
Online Accuracy  & $-0.06$ & $.572$ & $.992$ \\
Online F1        & $-0.54$ & $.162$ & $.844$ \\
Offline Accuracy & $-0.07$ & $.278$ & $.977$ \\
Offline F1       & $-1.51$ & $.019$ & $.885$ \\
\hline
\end{tabular}
\caption{Probe-by-band interactions in percentage points. Setting
$q$ values use BH-FDR-corrected sign tests over the prespecified
setting grid; model-cluster $q$ values use exact sign-flip tests over
the eight model means. The offline F1 interaction appears under the
setting-level analysis but is not retained after model clustering.}
\label{tab:probe_band_interactions}
\end{table}

The source-trained probes have lower unmodified target-test accuracy
than the target-trained probes (approximately $71.9\%$ versus $88.9\%$
online, and $70.7\%$ versus $88.9\%$ offline), confirming a meaningful
cross-dataset readout cost. Nevertheless, both own and transferred
bands remain disruptive under the source-trained probe. More
importantly, the own--transfer difference is small under both probes,
and none of the probe-by-band interactions is retained by the
model-clustered analysis.

These results separate two forms of transfer. Transferring the linear
readout reduces absolute target-domain performance, showing that the
complete decision boundary is not dataset invariant. In contrast, the
relative similarity between own and transferred layer bands remains
stable across probe choices. The main layer-transfer result is
therefore not an artifact of keeping the target-trained logistic
regression fixed. This supports partial sharing of emotion-sensitive
layer locations while also indicating dataset-specific readout
geometry.

Unrelated-band comparisons are computed only within a given probe.
We do not use them for cross-probe interactions when the source- and
target-probe runs contain different randomly selected unrelated bands.

\end{document}